\documentclass[conference]{IEEEtran}
\IEEEoverridecommandlockouts

\usepackage{amsmath,amssymb}
\usepackage{cite}
\usepackage{url}
\usepackage{booktabs}
\usepackage{array}
\usepackage{xcolor}
\usepackage{tikz}
\usetikzlibrary{arrows.meta,positioning,fit,backgrounds}

\begin{document}

\title{OptoAgent: A Trustworthy Multi-Agent Framework for Opportunistic Vision Micro-Screening in Classroom Environments}

\author{
	\IEEEauthorblockN{
		Toqeer Ali Syed\textsuperscript{1},
		Ali Akarma\textsuperscript{1,2}*,
		Adeel Ahmad\textsuperscript{1} and
		Hammad Muneer\textsuperscript{3}
	}
	\IEEEauthorblockA{
		\textsuperscript{1}AI Center, Faculty of Computer and Information Systems,
		Islamic University of Madinah, Madinah, Saudi Arabia\\
		\texttt{toqeer@iu.edu.sa} \quad
		\texttt{443059463@stu.iu.edu.sa} \quad \\
		\texttt{443057803@stu.iu.edu.sa}\\
		\textsuperscript{2}AI V\&V Lab, King Fahd University of Petroleum and Minerals,
		Dhahran, Saudi Arabia\\
		\textsuperscript{3}Department of Computer Science, The Islamia University of Bahawalpur, Bahawalpur, Pakistan\\
		\texttt{s21bdocs1m01009@iub.edu.pk}\\
		*Corresponding author: \texttt{443059463@stu.iu.edu.sa}
	}
}

\maketitle

\begin{abstract}
A child with reduced distance vision often does not know that anything is wrong.
Children adapt, move closer, and rarely report the difficulty, so the problem can
survive years of schooling before an adult notices. School screening addresses part
of this, but it runs on a schedule, depends on staffing, and is separated from the
classroom moments where the difficulty appears. Smartphone and web-based
acuity tests have widened access, yet every one of them still needs somebody to
start a test. We present SightSentinel, an architecture that turns a wall display a child already
reads from into a recurring screening site. Ordinary educational content carries
short calibrated optotype probes, and eight specialized agents divide the work.
Perception agents recover viewing distance, recognition accuracy, approach
behavior, gaze stability, response latency, and interocular difference from each
encounter. A quality agent discards observations taken under bad geometry, poor
lighting, or inattention. A longitudinal agent accumulates only the surviving
evidence against the child's own baseline, and an orchestrator reports a Vision
Concern Score routed through a safety gate whose output range excludes diagnosis,
refraction, prescription, and reassurance. The
design question is whether many cheap, noisy, well-gated encounters can reach a
referral decision that one scheduled test reaches late or misses. We state the
formulation, the architecture, a four-stage validation protocol against clinical
reference standards, and the conditions under which the approach should be rejected.
\end{abstract}


\section{Introduction}

Reduced distance vision in a child is easy to miss because the child has no
reference point for what normal vision looks like. Teachers and parents do see the
downstream signs, including squinting, drifting toward the board, losing their
place in text, and avoiding distant material, but each of those signs has many
innocent explanations and none of them carries a date stamp.

Screening programs exist to catch what informal observation misses. Al-Atawi's
synthesis of 35 studies covering more than 30,000 Saudi schoolchildren found that
refractive error and amblyopia remain common at school age while routine screening
coverage and parental awareness stay uneven \cite{AlAtawi2026}. Little et al.\ reach
a similar conclusion from the international literature and stress that referral and
follow-up matter as much as the screening event itself \cite{Little2025}. The
structural weakness is the same in both reviews: screening is an event, and the
functional difficulty is continuous.

Digital acuity testing has attacked the access side of that problem. A phone
application tested in 100 children agreed acceptably with conventional charts and
reached roughly 89\% sensitivity for subnormal acuity \cite{Raffa2022}. The DigiVis
paired-device web test has been validated in children aged 4 to 10 \cite{Allen2021},
and three years of service data from a pediatric ophthalmology service later showed
that most remote consultations produced usable acuity results \cite{Liu2026}. Custom
phone-based at-home testing has been evaluated in a large pediatric cohort
\cite{Roberts2026}, and an objective detector has been validated in 461 children
aged 12 to 47 months \cite{Du2026}.
Every one of these tools, however, begins when a person decides to run a test.

\textbf{The gap.} Two research lines sit next to each other without meeting.
Calibrated testing controls the stimulus but requires a deliberate session.
Behavioral work observes children continuously and controls nothing: the Hong Kong
Children Eye Study measured habitual reading distance in 2,363 children and linked
shorter distance to undercorrected myopia and poorer habitual vision
\cite{Yam2023}, which is a real signal but not one that can support a decision on
its own. What is missing is a system that keeps the calibrated stimulus and gives
up the deliberate session.

\textbf{Our position.} A wall display already knows the angular size of what it
shows. If it can also estimate how far away the child is standing and check whether
the child read a known symbol correctly, then a normal reading interaction becomes a
measurement. Any single measurement of that kind is weak. The claim we want to test
is that a long sequence of weak measurements, filtered hard for validity and scored
against the child's own history, supports a referral decision that a single annual
test reaches late.

We deliberately do not aim at diagnosis. Non-cycloplegic instrument refraction
underestimates refractive error in children, particularly hyperopia, and cannot
replace cycloplegic assessment \cite{NonCyclo2025}. A recent review of AI-enabled
pediatric ophthalmic systems reports growing capability alongside unresolved
questions about safety, workflow, and equity \cite{Somerville2026}. Both findings
point the same way: the appropriate output is a referral, not a label.

This paper contributes the following.

\begin{enumerate}
\item \textbf{A sequential formulation of screening.} We recast pediatric vision
screening as evidence accumulation over many low-cost encounters under an explicit
validity gate, rather than classification from one scheduled examination. We call
this Opportunistic Longitudinal Vision Screening (OLVS).

\item \textbf{An interaction model.} A Smart Vision Wall interleaves brief
calibrated optotype probes with the educational content a child is already reading,
so screening evidence arrives without a screening appointment.

\item \textbf{An agent decomposition with a bounded output range.} Eight agents
separate geometry, stimulus control, behavior, asymmetry, validity, history,
governance, and referral reasoning. An orchestrator fuses them into a Vision
Concern Score (VCS), and a Clinical Safety Gate constrains the output range so that
diagnosis, refraction, prescription, and reassurance are unreachable by
construction.

\item \textbf{A personal-baseline mechanism.} Quality-weighted temporal decay
against a within-child baseline makes an isolated poor encounter non-actionable,
which targets the false-positive burden that limits practical school screening.

\item \textbf{A validation protocol with stated failure conditions.} We specify four
evaluation stages against clinical reference standards and name the outcomes that
would refute the approach.
\end{enumerate}

Sections~\ref{sec:related} through~\ref{sec:arch} develop the position, the
formulation, and the architecture. Section~\ref{sec:eval} gives the protocol.
Nothing in this paper reports measured performance.

\section{Related Work and Positioning}
\label{sec:related}

\subsection{Screening programs and instrument-based screening}

School screening is the established route to finding children who would otherwise go
undetected, and the two recent reviews above agree on both its value and its
coverage limits \cite{AlAtawi2026,Little2025}. Instrument-based screening extends
reach to younger, preverbal, or developmentally delayed children because
photoscreening and autorefraction identify amblyopia risk factors without demanding
a symbolic response \cite{AAP2012}. Neither route changes the episodic character of
the encounter: a child who passes in March can develop or reveal a problem in
October.

\subsection{Digital and objective acuity measurement}

Self-administered and automated acuity testing is now credible in children
\cite{Raffa2022,Allen2021,Liu2026,Roberts2026,Du2026}, though a 2023 systematic
review of pragmatic trials found results heterogeneous enough to warrant more
real-world validation before wide deployment \cite{SelfVA2023}. These systems
demonstrate that the calibrated part of our design is achievable on commodity
hardware. They also share the trigger problem, because a test that nobody starts
produces no data.

\subsection{Naturalistic reading signals}

Habitual reading distance carries measurable information about visual status
\cite{Yam2023}, and eye-tracking work converts natural reading into fixation
duration, regression counts, and gaze stability features for automated assessment of
reading ability \cite{Hariyama2026}. The target variable there is reading skill
rather than acuity, which is precisely why we treat such features as priors on
validity and concern rather than as evidence of refractive error. Children move
closer for attention, posture, glare, and text difficulty.

\subsection{Agentic AI and multi-agent coordination}

Recent paradigms in agentic artificial intelligence demonstrate that decomposing
complex, multi-objective domains into specialized, collaborating autonomous agents
improves modularity, auditability, and operational robustness. Multi-agent
architectures have been successfully formulated for constraint-driven personal
planning, including joint personal finance and nutrition optimization under budget
and health constraints \cite{syed2025finagent,Syed2026FinNutriAgent}. In assistive
and inclusive technologies, agentic frameworks have been deployed to coordinate
multilingual disability-inclusive workflows \cite{syed2026fedagent} and to provide
adaptive personalized support for individuals with physical disabilities and
neurodivergence \cite{Siddiqui2026ADAPT}. In broader cyber-physical settings, agentic
coordination and digital twin frameworks manage complex urban and civil
infrastructure through autonomous, auditable decision workflows
\cite{syed2026climate,syed2026agenticdt}, while collaborative interactive
environments explore spatial interaction with preserved data integrity
\cite{syed2022car}.

However, existing agentic systems predominantly operate within conversational
interfaces, virtual task spaces, or macroscopic civil infrastructure. They have not
addressed the problem of micro-calibrated, opportunistic sensory screening embedded
within everyday physical learning environments. SightSentinel translates the agentic
paradigm to pediatric health triage by distributing sensory geometry, stimulus
calibration, behavioral priors, validity filtering, and governance across specialized
agents operating under a strictly bounded safety gate.

\subsection{What this leaves open}

Calibrated tools need an appointment. Behavioral studies avoid the appointment but
abandon stimulus control. AI ophthalmic systems mostly consume clinical images or
dedicated screening data \cite{Somerville2026}. And while agentic architectures have
advanced in personal planning \cite{syed2025finagent,Syed2026FinNutriAgent}, disability
inclusion \cite{syed2026fedagent,Siddiqui2026ADAPT}, and infrastructure monitoring
\cite{syed2026climate,syed2026agenticdt}, they have not been applied to ambient
physiological sensing. SightSentinel occupies the untaken combination, keeping
controlled micro-stimuli inside uncontrolled everyday behavior, and paying for the
resulting noise with aggressive validity filtering, multi-agent arbitration, and
repetition over time.

\section{Problem Formulation}
\label{sec:formulation}

\subsection{Encounter model}

Child $i$ interacts with an instrumented display at encounter times
$t = 1,\dots,T$. Each encounter yields a privacy-limited feature vector

\begin{equation}
\mathbf{x}_{i,t} =
[\,a_{i,t}, d_{i,t}, h_{i,t}, g_{i,t}, r_{i,t}, q_{i,t}, l_{i,t}\,],
\label{eq:features}
\end{equation}

collecting calibrated recognition accuracy $a$, viewing distance $d$, head pose and
approach behavior $h$, gaze-derived reading features $g$, response latency $r$,
test quality and confidence $q$, and display and lighting context $l$.
Reliably extracting head pose, gaze stability, and facial landmarks from ambient
video under unconstrained classroom illumination requires discriminative and
noise-resilient visual representations. In related domains processing
high-dimensional sensory and visual streams, hybrid deep learning architectures
that integrate convolutional neural networks for spatial feature extraction with
vision transformers and dimensionality reduction have demonstrated high classification
fidelity in noisy environments \cite{shaikh2024advancing}. SightSentinel adopts a
lightweight edge-deployed vision backbone to extract the geometric and behavioral
components of $\mathbf{x}_{i,t}$ on-device while immediately expiring raw image buffers.

Stimulus control follows from geometry. To subtend a visual angle $\theta$ at
measured distance $d$, an optotype of height $s$ satisfies

\begin{equation}
s = 2 d \tan\!\left(\frac{\theta}{2}\right),
\label{eq:geometry}
\end{equation}

so distance error propagates directly into an error in the tested acuity level. For
the small angles involved in acuity testing, $\theta \approx s/d$, and differentiating
gives

\begin{equation}
\frac{\delta \theta}{\theta} \approx \frac{\delta d}{d},
\label{eq:sensitivity}
\end{equation}

a one-to-one transfer of relative distance error into relative error in the angular
size actually presented. A 10\% distance error therefore mislabels the tested acuity
level by roughly 10\%, which is why $A_D$ suppresses an encounter outright once
distance uncertainty passes tolerance instead of reporting a wider interval.

Difficulty adapts by a staircase on the optotype level $z_t$,

\begin{equation}
z_{t+1} =
\begin{cases}
z_t - \Delta, & \text{correct response},\\
z_t + \Delta, & \text{incorrect response},
\end{cases}
\label{eq:staircase}
\end{equation}

and the system reports a screening interval rather than a point acuity, because the
probe count per encounter is small by design.

Where a validated separate-eye interaction is available under the deployment
protocol, the interocular difference

\begin{equation}
\Delta \mathrm{VA}_i = \left| \mathrm{VA}_{i,L} - \mathrm{VA}_{i,R} \right|
\label{eq:asym}
\end{equation}

between the left-eye and right-eye acuity estimates $\mathrm{VA}_{i,L}$ and
$\mathrm{VA}_{i,R}$ raises referral priority, since binocular viewing masks
unilateral loss.

\subsection{Validity before evidence}

The difference between an ambient screener and a surveillance heuristic is what it
refuses to score. An encounter contributes nothing unless

\begin{equation}
Q_{i,t} = Q_{\mathrm{dist}} \cdot Q_{\mathrm{light}} \cdot Q_{\mathrm{pose}}
\cdot Q_{\mathrm{attn}} \cdot Q_{\mathrm{disp}} \;\geq\; Q_{\min}.
\label{eq:quality}
\end{equation}

The product form is deliberate. Any single failed condition drives $Q_{i,t}$ toward
zero, so a wrong answer from a distracted child standing at an uncertain distance
cannot raise concern about that child's vision.

\subsection{Accumulation against a personal baseline}

Surviving evidence accumulates with temporal decay,

\begin{equation}
E_i(T) = \sum_{t=1}^{T} \lambda^{\,T-t}\, Q_{i,t}\, e_{i,t},
\label{eq:evidence}
\end{equation}

where $e_{i,t}$ is the encounter's evidence contribution and $\lambda \in (0,1]$
sets how fast old encounters lose weight. Each child is compared against their own
record, which is what separates a genuine decline from a bad afternoon.

The orchestrator reports a Vision Concern Score,

\begin{align}
\mathrm{VCS}_i = &\; w_1 A_i + w_2 D_i + w_3 B_i + w_4 G_i + w_5 L_i \nonumber\\
        &+ w_6 \Delta \mathrm{VA}_i + w_7 \mathrm{Trend}_i - w_8 \mathrm{QPenalty}_i,
\label{eq:vcs}
\end{align}

over calibrated acuity concern $A$, distance-dependent performance $D$, repeated
behavioral concern $B$, gaze concern $G$, latency concern $L$, interocular
asymmetry, deterioration from the personal baseline $\mathrm{Trend}$, and a penalty
for uncertain evidence. Synthesizing disparate, potentially noisy signals across time
mirrors multi-criteria optimization in goal-driven agentic architectures, where
coordinating agents balance competing behavioral attributes against rigid personal
constraints, such as financial budgets versus nutritional targets in personalized
planning agents \cite{syed2025finagent,Syed2026FinNutriAgent}. In Eq.~\eqref{eq:vcs},
the orchestrator balances immediate behavioral observations against longitudinal trend
stability and quality penalties, ensuring that fleeting behavioral variance does not
trigger an inappropriate clinical escalation. The score maps onto one of four
referral bands,

\begin{equation}
\begin{aligned}
\{&\text{Insufficient Evidence},\ \text{Routine Recheck},\\
  &\text{Vision Screening Recommended},\ \text{Prompt Clinical Review}\}.
\end{aligned}
\label{eq:bands}
\end{equation}

No band names a condition, and no band clears a child. ``Insufficient Evidence'' is
a statement about the data, not about the eye.

\section{Architecture}
\label{sec:arch}

The agent set is
$\mathcal{A} = \{A_D, A_O, A_B, A_E, A_C, A_L, A_P, A_R\}$, summarized in
Table~\ref{tab:agents}, and the orchestrator resolves their candidate outputs into a
single decision,

\begin{equation}
I^{*} = A_{\mathrm{orch}}(I_D, I_O, I_B, I_E, I_C, I_L, I_P).
\label{eq:orch}
\end{equation}

\begin{figure*}[!t]
\centering
\includegraphics[width=\textwidth]{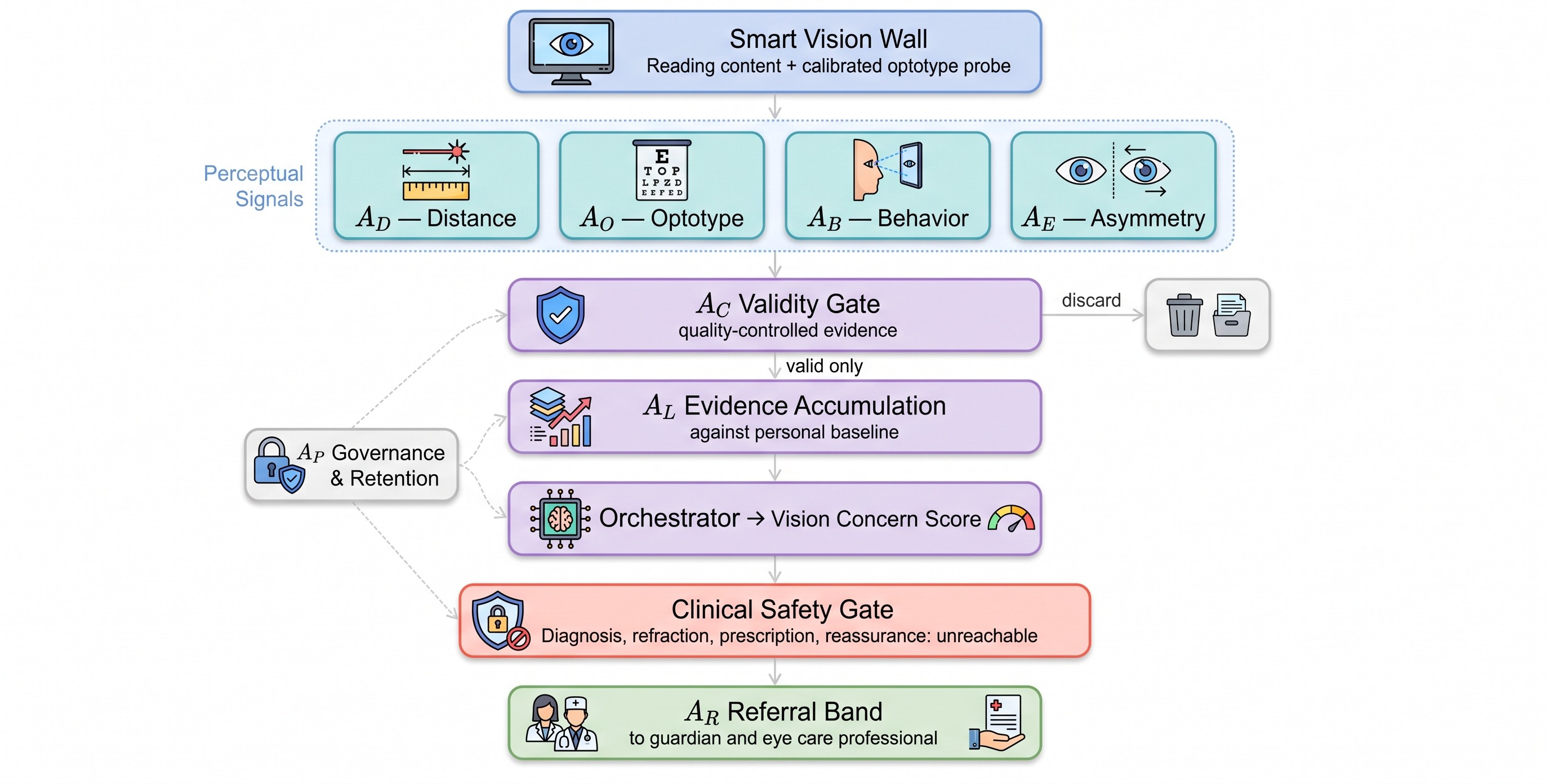}
\caption{Evidence path through SightSentinel. Perception agents run on every
encounter, the validity gate of Eq.~\eqref{eq:quality} discards encounters taken
under unusable conditions, and only surviving evidence reaches the accumulator. The
Clinical Safety Gate sits between the score and any human-facing output, so a
referral band is the only thing the system can say.}
\label{fig:arch}
\end{figure*}

Figure~\ref{fig:arch} shows the path evidence takes. The ordering carries the safety
argument. Because $A_C$ sits upstream of accumulation, a behavioral observation can
never outvote the fact that the observation was unusable, and because the safety
gate sits downstream of the score, no tuning of the weights in Eq.~\eqref{eq:vcs}
can produce a clinical claim.

\begin{table}[!t]
\caption{Agents and Their Responsibilities}
\label{tab:agents}
\centering
\renewcommand{\arraystretch}{1.15}
\begin{tabular}{@{}l l p{4.4cm}@{}}
\toprule
\textbf{Agent} & \textbf{Role} & \textbf{Responsibility} \\
\midrule
$A_D$ & Geometry   & Eye-to-display distance from depth, stereo, or calibrated
                     monocular cues. Suppresses the encounter when distance
                     uncertainty exceeds tolerance. \\
$A_O$ & Stimulus   & Age-appropriate optotype selection and staircase control.
                     Reports an acuity interval, never a point value. \\
$A_B$ & Behavior   & Approach, head tilt, squint-like narrowing, latency, gaze
                     stability, reading regressions. Weak priors only. \\
$A_E$ & Asymmetry  & Brief separate-eye screening where the protocol permits it.
                     Drives interocular priority. \\
$A_C$ & Validity   & Distance, lighting, pose, attention, and display checks.
                     Holds veto power over every encounter. \\
$A_L$ & History    & Decayed, quality-weighted accumulation against the child's
                     own baseline. \\
$A_P$ & Governance & On-device processing, buffer expiry, purpose limitation,
                     identity separation. \\
$A_R$ & Referral   & Maps the score to a band and routes it to the guardian
                     through an approved channel. \\
\bottomrule
\end{tabular}
\end{table}

\subsection{Interaction design}

The wall runs in two modes. Natural Reading Mode presents ordinary educational
material and yields only passive features, namely distance, pose, gaze, and approach
behavior. Micro-Screening Mode inserts a short calibrated challenge into the
existing activity, for example, identifying one symbol, letter, number, or picture
inside a classroom game, and adds the recognition accuracy term that anchors
Eq.~\eqref{eq:vcs}. Mode selection is itself a scheduling decision, since probing too
often turns a reading wall back into a test.

A single encounter contributes a handful of responses. That is the point: the burden
per encounter stays near zero, and statistical strength comes from
Eq.~\eqref{eq:evidence} instead.

\subsection{Worked example}

Consider a nine-year-old who handles a tablet at arm's length without complaint and
cannot resolve the same size of print once it appears on the board across the room.
Nothing about this strikes the child as unusual, so nobody hears about it.
Over several weeks the wall records encounters, most of which $A_C$ discards for
inattention or uncertain geometry. Among the encounters that survive, the child
repeatedly misses calibrated symbols at the standard classroom distance and succeeds
on the same level after moving closer. $A_D$ confirms the change tracks a real
distance change rather than a geometry error, and $A_L$ finds the pattern repeating
rather than isolated, so $\mathrm{Trend}_i$ and $D_i$ in Eq.~\eqref{eq:vcs} both
rise while $\mathrm{QPenalty}_i$ stays low.

The system does not report myopia, and it does not report a dioptric value. It
reports repeated distance-vision concern and recommends a formal visual-acuity and
eye examination. An optometrist or ophthalmologist then decides what is actually
wrong, which is the only point in the pathway where a diagnosis exists.

\subsection{The safety gate}

The gate is a constraint on the output range, written as

\begin{equation}
\text{SightSentinel} \not\rightarrow
\{\text{Diagnosis},\ \text{Prescription},\ \text{Treatment}\},
\label{eq:gate}
\end{equation}

with the sanctioned pathway

\begin{equation}
\begin{aligned}
\text{Persistent concern} &\rightarrow \text{formal screening}\\
&\rightarrow \text{eye care professional} \rightarrow \text{diagnosis}.
\end{aligned}
\label{eq:path}
\end{equation}

The system never emits ``myopia,'' ``amblyopia,'' a dioptric value, or ``eyes
normal.'' Confirmation stays clinical, and cycloplegic refraction remains the
reference where accommodation would bias a non-cycloplegic estimate
\cite{NonCyclo2025}. Prevent Blindness guidance likewise treats the evidence for
some machine-based school screening as insufficient, which argues for conservative
thresholds and prospective validation rather than early autonomy
\cite{PreventBlindness}.
Bounding the operational and semantic scope of autonomous agents is a fundamental
tenet of trustworthy agentic AI. As established by Syed et al., unconstrained
multi-agent pipelines and multimodal systems are vulnerable to goal drift, semantic
over-reach, and input manipulation unless protected by strict boundary enforcement
and defensive validation layers \cite{syed2025toward}. In SightSentinel, where the
inputs originate from uncurated classroom interactions, the Clinical Safety Gate
provides an architectural guarantee: by hardcoding Eq.~\eqref{eq:gate}, the system
eliminates the possibility that unexpected sensory inputs or agent deliberation drifts
could generate unauthorized diagnostic assertions or false clinical clearances.

\subsection{Privacy by architecture}

Continuous observation of children is the part of this design most likely to fail on
governance grounds, so the constraints are structural rather than procedural. Video
is processed locally where the hardware allows, raw frames live in a short rolling
buffer that expires after feature extraction, and the durable record holds
calibrated screening features rather than classroom footage. Identity is stored
apart from analytic features. Nothing resembling a medical inference reaches a
teacher. Referral wording travels to a guardian over a sanctioned route and nowhere
else, and $A_P$ enforces

\begin{equation}
\mathrm{DataUse} \subseteq \mathrm{Purpose}_{\mathrm{VisionScreening}},
\label{eq:purpose}
\end{equation}

which rules out attention scoring, behavior monitoring, academic profiling, and
advertising derived from screening data.
This structural posture operationalizes the core tenets of ethical AI governance
frameworks developed for sensitive IoT and connected sensing environments
\cite{jan2026eagf}. Under the four-pillar governance principles articulated by Jan et
al., encompassing privacy preservation, robust cybersecurity, accountability, and
ethical transparency, $A_P$ ensures that sensor telemetry remains purpose-bound and
cryptographically insulated. By enforcing Eq.~\eqref{eq:purpose} through on-device
ephemeral memory and role-restricted dissemination channels, the system prevents
secondary function creep and provides verifiable ethical governance in public
educational spaces.

\subsection{Distributed checkpoints}

Instrumented surfaces can sit at classroom boards, library walls, corridors, home
study areas, pediatric waiting rooms, and community kiosks. Under guardian consent
and secure identity matching, evidence from $N_c$ checkpoints merges as

\begin{equation}
E_i = \bigcup_{c=1}^{N_c} E_{i,c},
\label{eq:union}
\end{equation}

which raises encounter frequency without lengthening any individual interaction.
Aggregating micro-screening evidence across heterogeneous, distributed surfaces
mirrors the design of agentic digital twins and secure federated multi-agent
networks deployed in smart community and urban environments
\cite{syed2026climate,syed2026agenticdt,syed2026fedagent}. In such distributed
systems, edge-situated agents execute local validation, maintain temporal state, and
securely communicate decentralized observations to an orchestrating node without
exposing raw sensory streams to a central repository. Furthermore, maintaining
perceptual and geometric consistency across diverse interactive checkpoints draws on
principles established in collaborative augmented reality frameworks \cite{syed2022car},
guaranteeing that stimulus geometry, display calibration, and interaction logs maintain
data integrity across school and community environments.

\section{Evaluation Protocol}
\label{sec:eval}

The framework is a design. We have run no experiments, and the statements below
describe how it should be tested rather than what it achieved.

Four questions organize the work. Can calibrated probes embedded in ordinary reading
detect reduced distance acuity at useful sensitivity and specificity? Does fusing
many encounters beat a single ambient encounter? Do distance, approach, gaze, and
latency add anything beyond optotype performance alone? Can a within-child baseline
cut the false-positive referrals that attention, fatigue, and posture would otherwise
generate? Each maps to one stage below.

\textbf{Stage 1, calibration against clinical reference.} Recruit children spanning
a range of clinically measured acuity and refractive status, and compare
display-based micro-screening against standardized monocular acuity and clinical
refraction. Report the rejection rate of the validity gate, mean absolute error
against the reference acuity, test-retest reliability, and the sensitivity,
specificity, and area under the receiver operating characteristic curve for reduced
acuity. The rejection rate decides feasibility, since a system that discards most
of what it sees needs an implausible encounter rate.

\textbf{Stage 2, ablation of the fusion model.} Compare optotype performance
alone against optotype with distance, with behavior, with gaze, the full multimodal
model, and the longitudinal model. The informative outcome is the one that refutes
us: if optotype performance alone matches the full model, then $A_B$, $A_E$, and the
fusion weights of Eq.~\eqref{eq:vcs} should be removed rather than defended.

\textbf{Stage 3, prospective school deployment.} Run the system in consented
classrooms without disturbing normal teaching, and compare its referrals against
conventional school screening followed by clinical examination. The clinical
outcomes are previously unknown problems found, clinically relevant cases missed,
false-positive referral rate, and elapsed time to referral. The operational outcomes
are screening coverage, cost per confirmed referral, added teacher workload,
cooperation from the children themselves, and the privacy objections families raise
once the system is running rather than once it is described.

\textbf{Stage 4, longitudinal test of the central claim.} Measure the lead time
between the moment repeated ambient encounters raise a band and the moment a yearly
or sporadic screening round would have caught the same decline. OLVS lives or dies
here, and a null result refutes the paradigm regardless of how well Stages 1 to 3
go.

Accuracy is not the only thing that decides whether the system deploys. Two further
questions run alongside the stages above. The first is how much screening
performance on-device feature extraction with raw-imagery deletion costs relative to
a permissive-retention configuration, since a privacy design that destroys utility
gets switched off. The second is whether continuous ambient screening is acceptable
to the people living with it. Answering that needs a structured stakeholder study
across children, their families, classroom teachers, school administrators, and the
eye care professionals who receive the referrals, rather than a single metric.

\section{Discussion and Limitations}

Reduced performance at a wall display has many causes. Attention, literacy,
language, neurodevelopmental difference, fatigue, glare, ocular disease, and
refractive error all produce missed symbols, and behavior observed in a classroom
cannot separate them. Formally,

\begin{equation}
\text{ScreenPositive} \neq \text{Diagnosis}.
\label{eq:notdx}
\end{equation}

Three risks deserve naming. Validity gating is the load-bearing component and also
the least proven: if $Q_{i,t}$ is miscalibrated, the system either discards
everything or accumulates noise, and only Stage 1 data can settle that. The
asymmetry agent $A_E$ requires safe, supervised occlusion, which rules out
improvised procedures with unsupervised young children. Equity is a live concern,
because a system that works well at one reading level, language, or
optotype set and poorly at another would redistribute screening access rather than
widen it, which matches the implementation gaps reported for pediatric ophthalmic
AI \cite{Somerville2026}. Validation must therefore be stratified along every axis
that shifts the measurement, covering refractive status and age on the child's side,
reading language and optotype set on the stimulus side, and display size, ambient
lighting, and viewing distance on the environment's side.
In particular, children with neurodevelopmental differences (such as autism spectrum
disorder or ADHD) and those with motor or communication disabilities may naturally
exhibit idiosyncratic visual behaviors, such as intermittent fixations, prolonged
response latencies, or irregular approach dynamics. Recent developments in agentic
frameworks for people with disabilities and neurodivergence emphasize the necessity
of adaptive interaction profiles, multimodal expression channels, and barrier-aware
persona modeling \cite{Siddiqui2026ADAPT,syed2026fedagent}. Without such inclusive
accommodations, an ambient screener risks misclassifying neurodivergent behavioral traits
as indicators of visual deterioration, thereby inflating false-positive referrals.
Incorporating neurodiversity-aware baseline adjustments and non-verbal response modalities
will be essential in preventing disparate impact.

Finally, as with any connected sensing infrastructure deployed across distributed
educational facilities, system security is paramount. The communication links
connecting distributed checkpoints must be fortified against denial-of-service
disruptions, sensor spoofing, and malicious tampering, leveraging hybrid deep learning
threat detection and rigorous cyber-physical safeguards \cite{shaikh2024advancing,jan2026eagf}
to ensure that triage records remain authentic and uninterrupted.

The framework complements established screening pathways and does not substitute for
professional eye care.

\section{Conclusion}

Children adapt to poor vision quietly, and scheduled screening is not built to
notice quiet adaptation. SightSentinel responds by changing when screening happens
rather than how acuity is measured: calibrated probes ride inside ordinary reading,
a validity gate throws out most of what it sees, and evidence accumulates against
each child's own record until a referral is justified. The output stays a referral
band by construction, so the system can be wrong about concern without ever being
wrong about a diagnosis.

The paradigm stands or falls on one empirical question: whether many weak gated
encounters detect deterioration earlier than an infrequent strong test. We
have specified the protocol that answers it, including the results that would tell
us to abandon the approach. Until those studies run, SightSentinel is an
architecture and a hypothesis.

\bibliographystyle{IEEEtran}
\bibliography{references}

\end{document}